\documentclass[fleqn,10pt]{wlscirep}
\usepackage[utf8]{inputenc}
\usepackage[T1]{fontenc}
\usepackage{bm}
\usepackage{amsmath,amsfonts,amscd,amssymb}
\usepackage{xcolor}
\usepackage{graphicx}
\usepackage{epstopdf}
\usepackage{overpic}
\usepackage{cancel}
\usepackage{rotating}
\usepackage{url}
\usepackage{caption}
\usepackage{color}
\usepackage{rotating}
\usepackage{multirow}
\usepackage{wrapfig}
\usepackage{mathtools}
\usepackage{subeqnarray}
\usepackage{setspace}
\usepackage{palatino} % needs 10
\usepackage[numbers,sort&compress]{natbib}

\usepackage[bottom,flushmargin,hang,multiple]{footmisc}
\usepackage{lipsum}

\definecolor{header1}{cmyk}{0,0,0,1}

\usepackage[utf8]{inputenc}

\usepackage[normalem]{ulem}
\usepackage{color}
\title{Decoding complexity: how machine learning is redefining scientific discovery }

\author[1,2*]{Ricardo Vinuesa}
\author[3]{Paola Cinnella}
\author[4]{Jean Rabault}
\author[5,2]{Hossein Azizpour}
\author[6,7]{Stefan Bauer}
\author[8]{Bingni W. Brunton}
\author[9,2]{Arne Elofsson}
\author[10,2]{Elias Jarlebring}
\author[5,2]{Hedvig Kjellstr\"om}
\author[11,2]{Stefano Markidis}
\author[12,13]{David Marlevi}
\author[14*]{Javier Garc\'ia-Mart\'inez}
\author[15*]{Steven L. Brunton}

\affil[1]{FLOW, Engineering Mechanics, KTH Royal Institute of Technology, Stockholm, Sweden}
\affil[2]{Swedish e-Science Research Centre, (SeRC), Stockholm, Sweden}
\affil[3]{Institut Jean le Rond D'Alembert, Sorbonne Université, France}
\affil[4]{IT Department, Norwegian Meteorological Institute, 0313 Oslo, Norway}
\affil[5]{Robotics, Perception and Learning, KTH Royal Institute of Technology, Stockholm, Sweden}
\affil[6]{TUM School of Computation, Information and Technology, Technical University Munich, Munich, Germany}
\affil[7]{Helmholtz AI, Helmholtz Center Munich, Munich, Germany}
\affil[8]{Department of Biology, University of Washington, Seattle, WA 98195, USA}
\affil[9]{Dept. of Biochemistry and Biophysics and Science for Life Laboratory, Stockholm University, 171 21 Solna}
\affil[10]{Dept. Mathematics, KTH Royal Institute of Technology, 100 44 Stockholm, Sweden }
\affil[11]{Department of Computer Science, KTH Royal Institute of Technology, Stockholm, Sweden}
\affil[12]{Dept. Molecular Medicine and Surgery, Karolinska Institutet, 171 77 Stockholm, Sweden}
\affil[13]{Inst. for Medical Engineering and Science, Massachusetts Institute of Technology, Cambridge, MA 02139, USA}
\affil[14]{Departamento de Qu\'imica Inorg\'anica, Universidad de Alicante, Alicante, Spain}
\affil[15]{Department of Mechanical Engineering, University of Washington, Seattle, WA 98195, USA}
\affil[*]{E-mails for correspondence: rvinuesa@mech.kth.se, j.garcia@ua.es, sbrunton@uw.edu}

\vspace{-.2in}

\begin{abstract}
As modern scientific instruments generate vast amounts of data and the volume of information in the scientific literature continues to grow, machine learning (ML) has become an essential tool for organising, analysing, and interpreting these complex datasets. This paper explores the transformative role of ML in accelerating breakthroughs across a range of scientific disciplines. By presenting key examples -- such as brain mapping and exoplanet detection -- we demonstrate how ML is reshaping scientific research. We also explore different scenarios where different levels of knowledge of the underlying phenomenon are available, identifying strategies to overcome limitations and unlock the full potential of ML. Despite its advances, the growing reliance on ML poses challenges for research applications and rigorous validation of discoveries. We argue that even with these challenges, ML is poised to disrupt traditional methodologies and advance the boundaries of knowledge by enabling researchers to tackle increasingly complex problems. 
Thus, the scientific community can move beyond the necessary traditional oversimplifications to embrace the full complexity of natural systems, ultimately paving the way for interdisciplinary breakthroughs and innovative solutions to humanity's most pressing challenges.

\noindent\emph{Keywords:} machine learning (ML); deep learning (DL); artificial intelligence (AI); scientific discovery; complexity; physics; life sciences; computer science

\end{abstract}
\begin{document}
\flushbottom
\maketitle

\thispagestyle{empty}

\section*{Introduction}\label{sec:intro}
Machines have played a critical role in scientific discovery ({\i.e.} to obtain fundamental and formalized knowledge about Nature) by providing the tools to observe, measure, and analyze natural phenomena. Scientific instruments, such as telescopes and microscopes, have historically enabled groundbreaking discoveries by revealing details invisible to the naked eye, expanding our understanding of the universe and the microscopic world~\cite{j1}. With the advent of modern scientific instruments, including DNA sequencers, astronomical observatories, and high-resolution imaging devices, research facilities are producing terabytes or even petabytes of information. As data volumes grow, computers play a critical role in organizing, analyzing, and interpreting this information. Advanced computational methods help to reduce the complexity of the data, making it possible to extract meaningful insights~\cite{j4,j5}. However, even with the most advanced computers, the sheer volume of data generated by large-scale projects such as the Large Hadron Collider (LHC) and the Square Kilometre Array (SKA), and the vast amount of information available in the scientific literature, make traditional analysis methods impractical. Complex problems such as weather forecasting, drug discovery, and genomic analysis often involve highly complex data sets and processes that cannot be efficiently managed without the assistance of machine learning (ML), which can help sift through massive data streams, identify patterns, and extract valuable insights that would be impossible for humans or traditional computational methods alone. Complexity in these problems arises from non-linearity, high dimensionality, and multiscale dynamics, posing significant challenges for traditional mathematical tools and even recent simulation-based approaches. Despite advancements in simulations and big data, we still struggle to fully understand phenomena like turbulence or biological processes at a deeper level. Advanced ML tools are also improving decision-making, enabling faster and more accurate interpretations of complex phenomena, and addressing challenges in diverse scientific fields~\cite{j6}. However, it also introduces some challenges and the need for ethical guidelines to ensure the appropriate use of ML for scientific research~\cite{j7}. A key issue is algorithmic bias, which can distort outcomes and lead to incorrect conclusions, particularly in areas like health care, where biased predictions can impact patient care or drug development. Additionally, the black-box nature of ML models complicates transparency, making it hard for researchers to verify decisions. Lastly, data ownership and privacy concerns arise, especially with sensitive data like genetic or health records. These issues require the development of robust ethical guidelines to ensure that ML contributes to science in a fair, transparent, and socially responsible way~\cite{ethics_ref}.

\begin{figure}
\centering 
\includegraphics[width=0.75\textwidth]{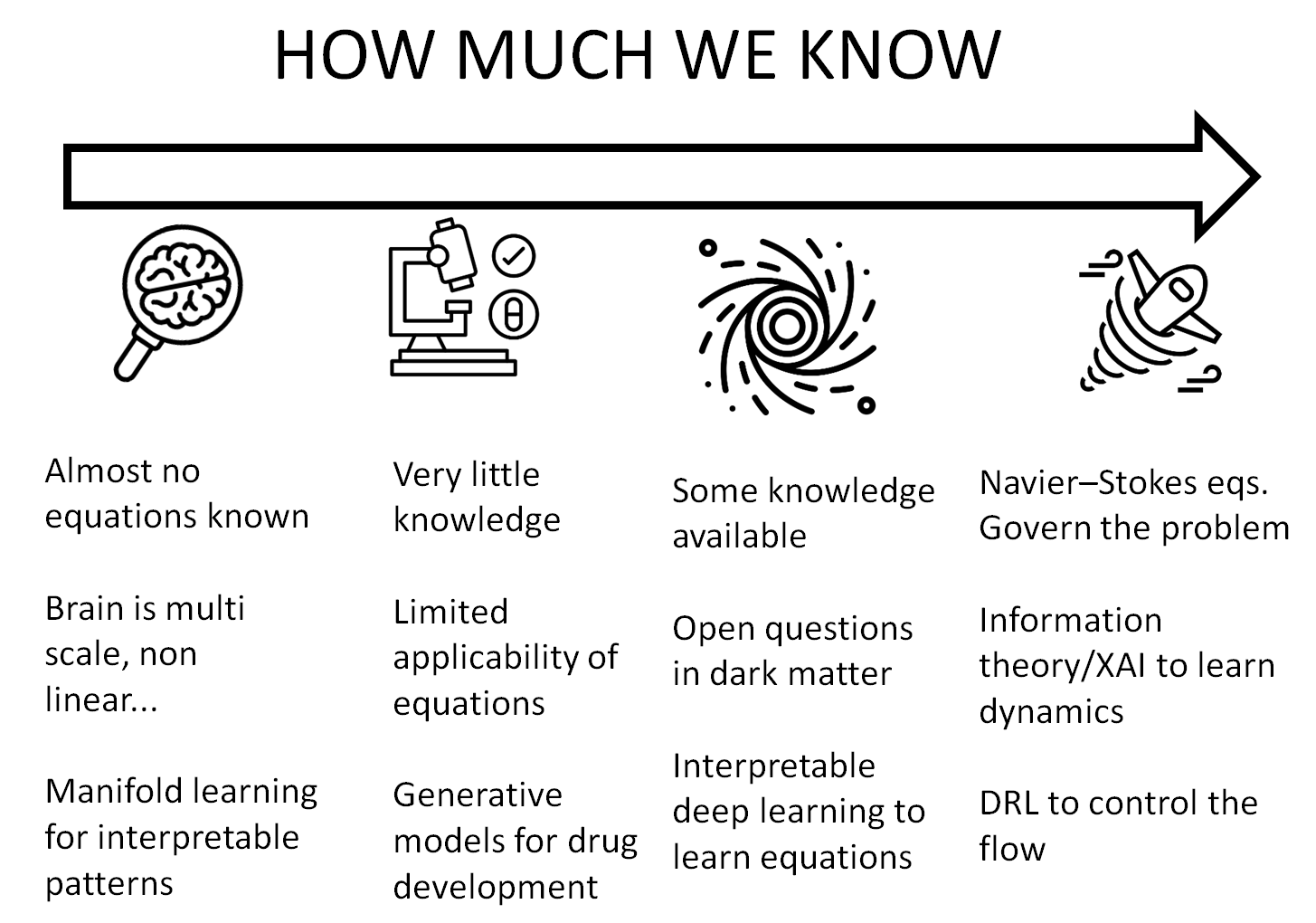}
   \caption{{\bf Schematic representation of the various applications of ML for scientific discovery, depending on the amount of knowledge available in each category.} A number of examples are provided, including brain research, drug discovery, dark matter and fluid mechanics.}
   \label{fig:summary_intro}
\end{figure}

In this work, we explore the potential of ML and artificial intelligence (AI) in three types of scientific problems: (i) those where all governing equations are known, (ii) those with partial knowledge and (iii) those where little is  known. We illustrate this with examples from the physical and life sciences, including turbulent flows, dark matter, drug discovery, and brain research. Figure~\ref{fig:summary_intro} summarizes how different uses of ML help address complexity in these areas. As discussed below, when the system complexity grows, the distinction between full, partial, and no knowledge of the governing equations becomes increasingly blurred, but ML plays a crucial role in tackling these challenges. A good  example of a  complex system with vast amounts of data that traditional tools cannot process efficiently is brain research. ML enables the reconstruction of countless brain slices into highly accurate three-dimensional (3D) maps. In a recent study, Google researchers used AI to process 300 million brain images from Harvard, creating the largest-ever interactive 3D brain tissue model, now available online. This innovation is crucial for understanding neurological disorders, as ML can detect patterns that traditional methods might miss, supporting early diagnosis and treatment planning~\cite{j10}. One of the most widespread  applications of ML is in drug discovery, addressing the time and cost challenges of traditional methods. The drug-development process can take over a decade and billions of dollars due to the complexity of identifying viable candidates. ML is transforming this by rapidly analyzing vast biological and chemical data, uncovering patterns that might remain hidden, and streamlining the identification of promising candidates. Additionally, ML enhances predictive modeling, allowing researchers to forecast drug efficacy and safety earlier in the process. By enabling virtual screening of drug-target interactions, ML reduces the need for costly lab experiments and helps design more efficient clinical trials, accelerating development and optimizing resources~\cite{j12}. 

\begin{figure}[t]
\centering 
\includegraphics[width=0.65\textwidth]{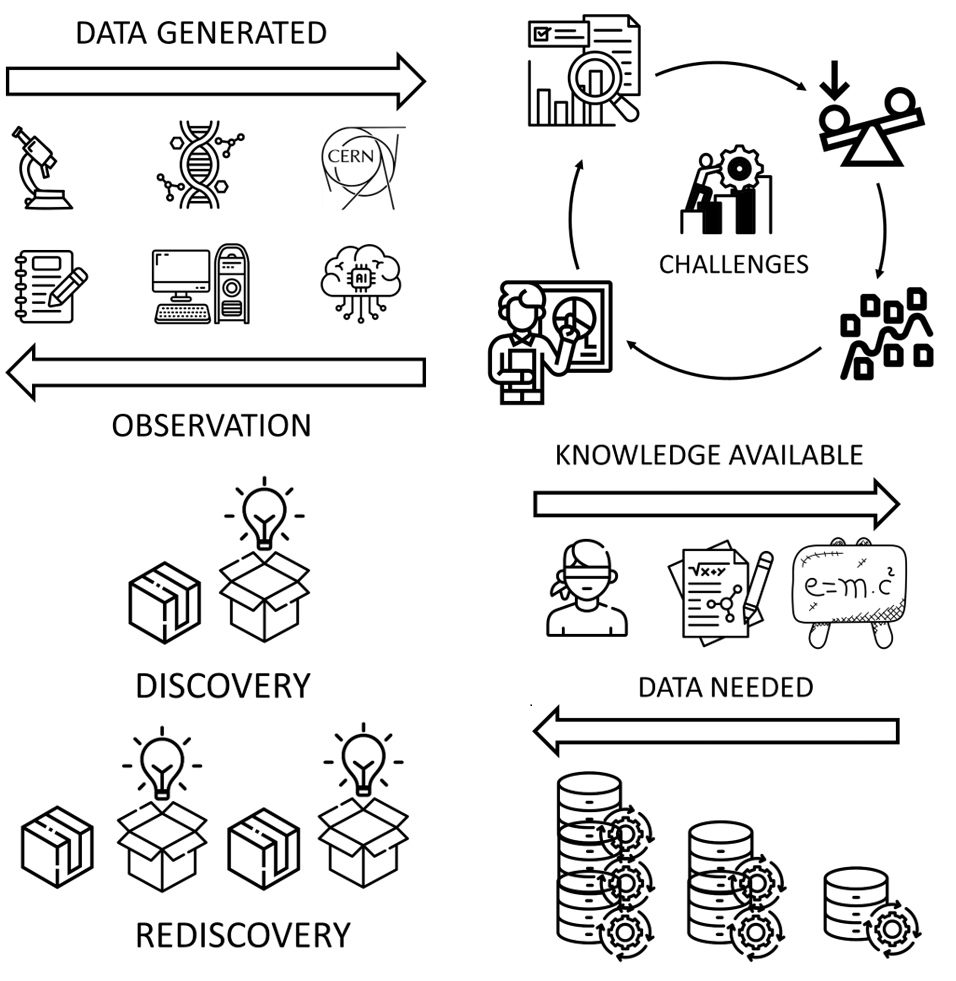}
   \caption{{\bf Visual summary of the paper illustrating some of its main ideas.} (Top left) The increase in the amount of data generated by scientific instrumentation over time and the shift from data organization by humans, computers, and finally by machine-learning techniques, which translates into less observation, intervention and understanding of humans on scientific discoveries. (Top right) The four challenges we have identified in the paper on the use of ML techniques for scientific discovery, these being: data quality and availability, potential biases, explainability and overfitting. (Bottom left) What constitutes a scientific discovery and the possibility of ML making original breakthroughs versus simply rediscovering known ideas, concepts or laws. (Bottom right)  The need to have more and better data to be able to make scientific discoveries with ML as we have less knowledge on the subject of study}
   \label{fig:conclusion}
\end{figure}

Figure~\ref{fig:conclusion} provides a comprehensive visualisation of the key themes discussed in this paper, focusing on the role of machine learning (ML) in scientific discovery and the challenges it presents, particularly with respect to data. The top-left panel illustrates the significant increase in data generated by modern scientific instrumentation over time. Initially, humans organised and interpreted the data through manual analysis and generalisation. Since then, computational methods have facilitated the management of large amounts of data, contributing significantly to fields such as genomics, chemistry and mathematics. More recently, with the advent of large-scale scientific facilities such as CERN, ML techniques have become essential for processing vast amounts of data. This shift has reduced human involvement in direct observation and interpretation, raising concerns about our diminishing understanding of the discoveries made by these technologies. The top-right panel outlines four key challenges related to ML in science: data quality and availability, inherent biases, explainability, and the risk of overfitting. The bottom-left panel highlights a conceptual dilemma: while ML accelerates discovery, there is growing debate about what constitutes a true scientific breakthrough, and whether ML can only ``rediscover'' existing concepts rather than uncover new insights. Finally, the bottom-right panel emphasises the increasing need for high-quality data to enable new discoveries through ML, while highlighting the limitations imposed by our current gaps in scientific knowledge. All of these issues are discussed in detail in the following sections of this paper. 

Artificial Intelligence is playing a transformative role in physics by improving data analysis, model development, and experimental interpretation. In astronomy, ML is improving the search for exoplanets by boosting the accuracy and efficiency of data analysis. AI-powered algorithms, particularly convolutional neural networks, can process massive data sets from telescopes to detect Earth-like exoplanets in noisy signals more precisely than traditional methods. The transit method, which detects exoplanets by observing mini-eclipses as they pass in front of their stars, can be complicated by planetary interactions that disrupt periodicity. To address this, researchers from the University of Geneva, University of Bern, and Disaitek applied ML and image recognition techniques to predict these interactions. By training a neural network on numerous examples, they built a model capable of detecting subtle exoplanet signals that might be missed by traditional methods. Their work led to the discovery of exoplanets Kepler-1705b and Kepler-1705c, advancing our understanding of planetary systems~\cite{j13}. ML is also crucial in areas like the Standard Model of particle physics~\cite{j13}. Automated algorithms were central to the discovery of the Higgs boson~\cite{j14}, and future experiments will need to operate at higher energies and intensities, generating data volumes too large for traditional methods to handle. For instance, it is expected that the Large-Hadron Collider (LHC) will increase proton collision rates by an order of magnitude in the next decade, requiring data analysis tools such as ML, to identify trends, uncover hidden relationships, and design more effective experiments.

Another scientific area in which ML is increasingly assisting research is mathematics, in this case by improving the process of theorem proving, mathematical method development and discovery. ML systems have already demonstrated their ability to automate aspects of theorem proving; for example, Meta AI's neural theorem prover successfully solved 10 International Math Olympiad (IMO) problems, far exceeding the performance of previous ML systems. Another notable example is DeepMind's collaboration with mathematicians, which has resulted in ML contributing to new mathematical methods in knot theory and representation theory. Specifically, in 2021, ML was used in a new constructive way to suggest mathematical proofs. A collaboration between mathematicians and DeepMind demonstrated that ML can complement human intuition in proving or suggesting complex theorems. The team used ML to investigate long-standing conjectures, including the Kazhdan--Lusztig polynomials, and discovered new connections in knot theory. This illustrates AI's potential to accelerate mathematical research and open new frontiers~\cite{j16}.

\section*{Embracing complexity}

Machine learning comprises a growing set of algorithms, enabled by high-performance computing and increasingly vast data, that show incredible promise for handling complexity~\cite{drl_nuclear,wong2024ai}. Neural networks, despite being governed by simple rules, can perform complex tasks for which no traditional algorithms exist. Although they are fully observable and deterministic, we often cannot explain their decisions. However, they have led to groundbreaking discoveries, such as a new class of antibiotics~\cite{Wong38123686}. This creates challenges, such as the need for explainable AI (XAI)~\cite{rudin,vinuesa_interp}. Symbolic approaches, such as gene-expression programming, sparse regression, and sparse Bayesian learning~\cite{ferreira2006gene,schmidt2009distilling,Brunton2016pnas}, have been successful, but their complexity grows exponentially with the search-space size. Recent efforts to combine symbolic and deep-learning approaches have enabled advances, such as discovering new materials~\cite{Merchant38030720}. This raises fundamental questions about the limits of ML in scientific discovery; for example, can a complex system understand its own complexity? And how much can AI discover beyond its training data~\cite{leslie2023does}? These issues highlight the opportunities and challenges that data-driven methods bring to science. Some believe in the ``unreasonable effectiveness of data''~\cite{halevy2009unreasonable}, particularly in deep learning~\cite{sejnowski2020unreasonable}, but the practical implications for future discoveries remain uncertain. A key question is whether ML can provide not only computational solutions but also fundamental scientific understanding. Note that ML's applications in science are not limited to discovery. For example, AI is revolutionizing optimization, laboratory automation, and solving governing equations, such as partial differential equations (PDEs). Recent studies~\cite{wang_et_al_nature} have focused on the potential of self-supervised learning in experiments and simulations (including representations of scientific data), while others~\cite{zenil_discovery} are exploring AI's role in formulating scientific questions. Note that the unique contribution of this work is the study of how ML can tackle complexity to reach scientific discoveries, acknowledging that different levels of knowledge of the governing equations (see Figure~\ref{fig:summary_intro}) will require completely different ML approaches. This may constitute a revolution in how we organize disciplines when using ML for scientific discovery, where apparently different communities may share many similarities in terms of how much is known regarding the governing equations and therefore in terms of the ML methods to be developed.

The emergence of scientific foundation models (SFMs) and large language models (LLMs) is further pushing the boundaries of ML methods. Foundation models are large machine-learning or deep-learning generative models trained on vast amounts of data so they can be applied on a wide range of cases. They predict masked data regions to learn associations, excelling in multiple tasks without large, labeled datasets. Despite risks like bias and transparency issues~\cite{van2023chatgpt}, foundation models have revolutionized AI, especially through chatbots like ChatGPT~\cite{ChatGPT}. LLMs now assist with tasks from writing and coding to guiding scientific experiments~\cite{Bran38799228,Boiko38123806} and generating ideas~\cite{wang2023scimon}. Foundation models also show promise in non-text-focused areas, such as protein-structure prediction~\cite{esmfold}, protein design~\cite{hsu2022learning} and climate simulations~\cite{chen2023foundation}. 
%%%
\textcolor{black}{An intriguing feature of foundation models, and particularly LLMs, is that they demostrate so-called "emergent abilities" \cite{wei2022emergent,schaeffer2024emergent}. The term refers to unexpected, not explicitly programmed, capabilities that arise as model scale increases, and do not consist in mere extrapolations of smaller models' performance. 
Such abilities were first observed in complex natural systems, e.g. phase changes in materials, complex behavior in flocks of birds or fishes, as well as in computational systems such as cellular automata and agent-based models, and the term was originally popularized by the Nobel-prize winner P.W. Anderson \cite{anderson1972more}. Examples of LLM emergent abilities include in-context learning, complex reasoning, and multi-step problem-solving, which are highly valuable for scientific research. Thank to this, LLMs are found to generalize to new tasks without prior examples (zero-shot learning) or with minimal data (few-shot learning), making them particularly useful in domains where data is scarce or highly specialized. More in general, they have the potential to aid hypothesis generation, automate literature reviews, predict protein structures, and accelerate scientific discovery, especially when combined with reinforcement learning (RL) and a suitable system of rewards/penalties to optimize specific scientific tasks, such as designing experiments or iteratively refining hypotheses based on feedback. Synergy with scientists may allow to leverage the models' reasoning capabilities in dynamic, real-world scenarios, enabling more efficient exploration of complex scientific problems.
On the other hand, these abilities are often unpredictable and non-linear, raising challenges in reliability, interpretability, and ethical use. Careful oversight is then needed to ensure these models align with scientific goals and produce trustworthy results. Risks include the potential for generating misleading or incorrect outputs, amplifying biases present in training data, and over-reliance on automated systems without sufficient human validation. Additionally, the opacity of how these models arrive at their conclusions can hinder their adoption in critical scientific applications, raising important questions about their benefits and risks in science~\cite{birhane2023science}}. Fajardo-Fontiveros {\it et al.}~\cite{fajardo2023fundamental} discuss when it is possible to learn models from data and the acceptable noise levels for accurate learning.

\section*{Discovery versus re-discovery by machine learning}

While ML has proven invaluable in refining existing knowledge, its real potential lies in detecting patterns and correlations that may not be immediately apparent due to the vast amounts of data involved. This raises the question of whether ML is truly discovering new insights or merely reinterpreting existing knowledge. Historically, scientific discovery has been rooted in inquiry, observation, and experimentation, with creativity playing a crucial role in generating new insights into phenomena or theories~\cite{j17,j21}. While ML excels at analyzing large datasets, identifying patterns, and generating hypotheses (which are key components of discovery), its ability to make truly autonomous, original discoveries is still debated. ML models operate without a genuine understanding of the underlying mechanisms they analyze, meaning that their ``discoveries'' tend to be instrumental rather than original. Groundbreaking discoveries usually require creativity, broader contextual understanding and sometimes leaps beyond available data, all of which ML currently lacks. Although ML is revolutionizing research by uncovering complex trends, the idea that it can independently produce original scientific discoveries remains controversial. Several studies have explored this question~\cite{j22}, and workshops, such as the one organized by the National Academies on How ML Is Shaping Scientific Discovery, have debated the issue~\cite{j25}. Nevertheless, ML has already contributed to significant scientific breakthroughs. In drug discovery, ML identified new antibiotics, such as halicin, which were unknown and can kill antibiotic-resistant bacteria~\cite{j26}. ML also plays a crucial role in materials science, predicting new materials for batteries and superconductors using physical and experimental data. These achievements highlight AI's potential to navigate uncharted scientific territory~\cite{j27}.

One of the most significant advancements in AI-driven scientific discovery is AI-Hilbert. Developed by IBM researchers, AI-Hilbert acts as an ``AI scientist'' that transforms existing theories and data into new, consistent mathematical models. Its goal is to accelerate scientific discovery by automating hypothesis generation and testing. AI-Hilbert helps scientists uncover new knowledge by analyzing large scientific datasets and revealing patterns overlooked by traditional methods. It also refines theories by managing conflicting data~\cite{j28}. Furthermore, Cornelio {\it et al.}~\cite{j29} have developed a new ML tool which combines axiomatic knowledge with experimental data to derive scientific models. By integrating logical reasoning and symbolic regression, it has rediscovered laws like Kepler’s third law and Einstein’s time-dilation law. This tool can distinguish between competing formulas, even with limited data. While efficient at replicating human discoveries, these tools often confirm known theories rather than offering new insights. ML can be biased towards existing patterns, limiting its ability to generate novel hypotheses. In these cases, AI’s role is more supportive, validating established knowledge rather than challenging it~\cite{j30}. ML holds immense potential for making groundbreaking discoveries, particularly in fields like drug development and astrophysics. However, its frequent rediscovery of known scientific principles illustrates the current limitations of its creativity. For ML to drive new knowledge, it must evolve beyond confirming human findings. Currently, ML lacks an intrinsic understanding of the mechanisms it analyzes, so its discoveries often complement human expertise, relying on human interpretation and creativity to fully appreciate and exploit the insights gained.

\section*{Machine-learning-driven scientific discovery when complete information is available}\label{sec:everything}

\begin{figure}[t]
\centering 
\includegraphics[width=0.9\textwidth,angle=0,origin=c]{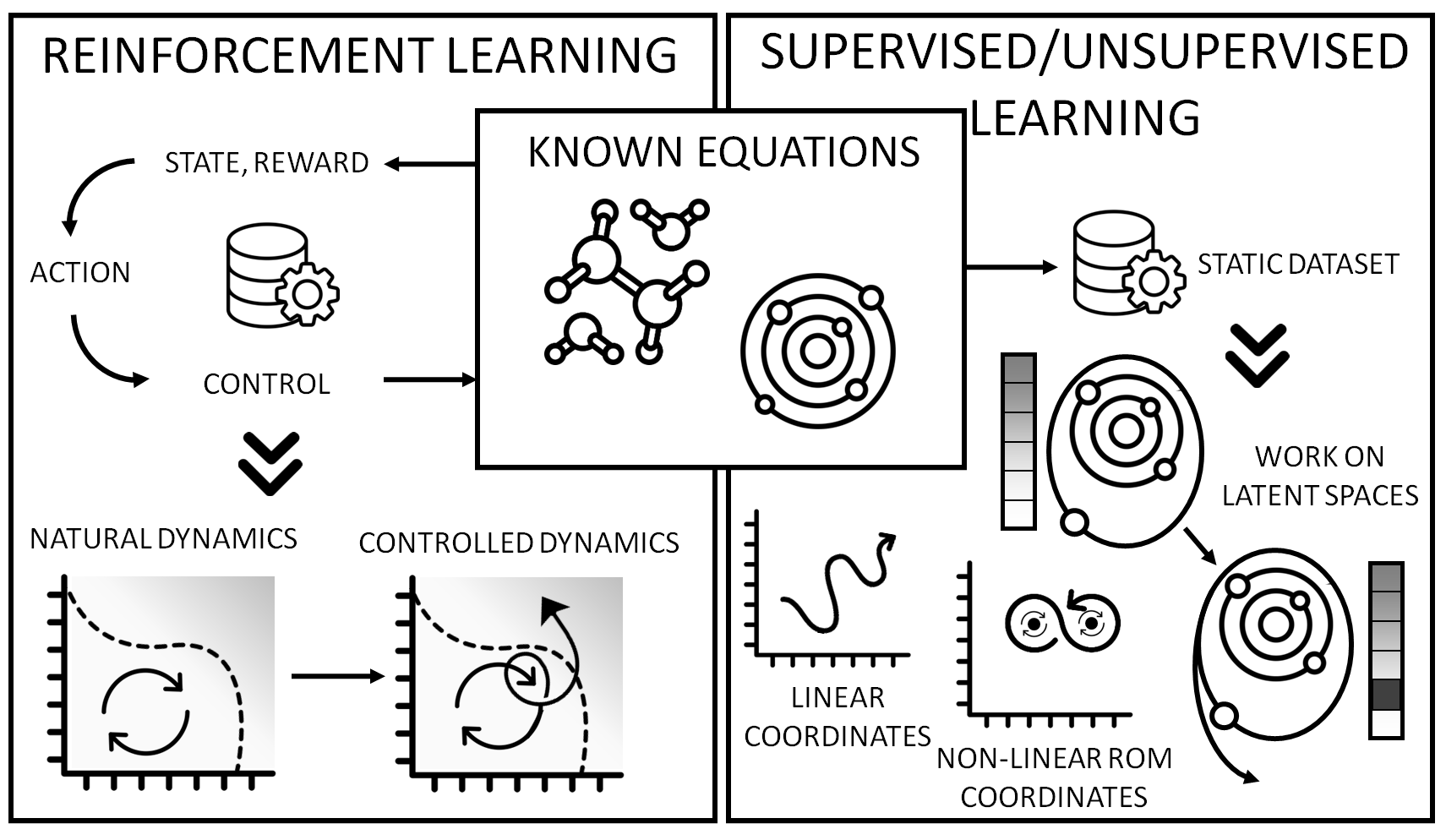}
   \caption{{\bf Schematic representation of ML directions to enable scientific discoveries when complete information about the governing equations is available.} In such a case, both supervised, unsupervised, and reinforcement-learning methodologies can be used. Supervised and unsupervised methodologies are made possible by \textcolor{black}{generating} large datasets of synthetic data simulated from the governing equations. This allows \textcolor{black}{the deployment of} a variety of ML techniques that can discover complex hidden relations, nonlinear coordinate systems, hidden dynamics or solve \textcolor{black}{otherwise intractable problems}. Reinforcement learning can also be used by coupling it to the physics simulator, which has already proven successful at discovering previously unknown control strategies and regimes of complex systems or \textcolor{black}{generating} high-quality heuristic guesses that can be tested in the case of problems where solution verification is easy, but the suggestion of good candidate solutions is hard.}
   \label{fig:know_everything}
\end{figure}

There are several applications within Physical Sciences where the underlying governing equations are known perfectly but the high-level global dynamics are still not well understood. For instance, while we know the quantum equations behind biomolecular dynamics, complexity makes biology a partial-knowledge problem since only limited biological systems can be measured or simulated, and simulating the full human brain is impossible. Similarly, turbulent flows, described by the Navier--Stokes equations, are only partially understood due to their chaotic nature as the Reynolds number increases~\cite{chapman1990mathematical}. While the large scales in the flow can be simulated, the smaller scales are much more difficult to simulate, making turbulence a major challenge~\cite{fefferman2000existence}. However, some ML techniques, such as symbolic regression and reduced-order modeling, can help uncover unknown flow features and physical properties from large direct-numerical-simulation (DNS) datasets~\cite{cremades_et_al,lozano_arranz}, as indicated in Figure~\ref{fig:know_everything}. Advances in turbulence modeling using supervised ML have also improved closure models for Reynolds-averaged Navier--Stokes (RANS) and large-eddy-simulation (LES) turbulence models~\cite{duraisamy_et_al,Brunton2020arfm}. In astrophysics, the supervised classifier SPOCK predicts long-term stability in multi-planet systems (which requires integration of the laws of gravitation over billions of orbital periods) using short-term simulations and effectively generalizing to larger systems~\cite{tamayo2020predicting}.

Optimal control of complex physical and biological systems is another challenging area where ML, particularly deep reinforcement learning (DRL), shows promise. DRL has led to breakthroughs in quantum physics, astronomy, turbulence control, and tokamak-instability control, offering insights into complex systems and discovering novel strategies~\cite{alpha_go}. DRL has even uncovered previously unknown thermodynamic cycles~\cite{beeler2021optimizing}. ML can dramatically accelerate simulations of complex systems with known equations. Autoencoders can be used to discover latent-space representations that enable faster time integrators and simulations, leading to better optimization and systematic studies~\cite{solera_rico,park2022optimization}. For instance, high-energy physics relies on comparing observed particle detector data with simulations, which are computationally intensive. Fast generative models like generative adversarial networks (GANs) and variational autoencoders (VAEs) offer a faster alternative, although on-going research aims at ensuring that the required accuracy is achieved~\cite{goodfellow_et_al,albertsson2018machine}. In climate science, foundation models for weather forecasting are revolutionizing climate studies, offering potential breakthroughs in understanding climate change and paleoclimates~\cite{wong2024ai}. In applied mathematics, DRL and large language models (LLMs) are generating novel algorithms and optimizations. Notable examples include matrix operations~\cite{alphatensor} and combinatorial problems, where AI aids in discovering new strategies that can be validated with classical algorithms~\cite{RomeraParedes2024Mathematical}. In these cases, AI provides valuable heuristic methods, although it must be highlighted that finding adequate solutions still remains challenging.

\section*{Machine-learning-driven scientific discovery when only partial information is available}\label{sec:something}

\begin{figure}[t]
    \centering
    \includegraphics[width=0.9\linewidth]{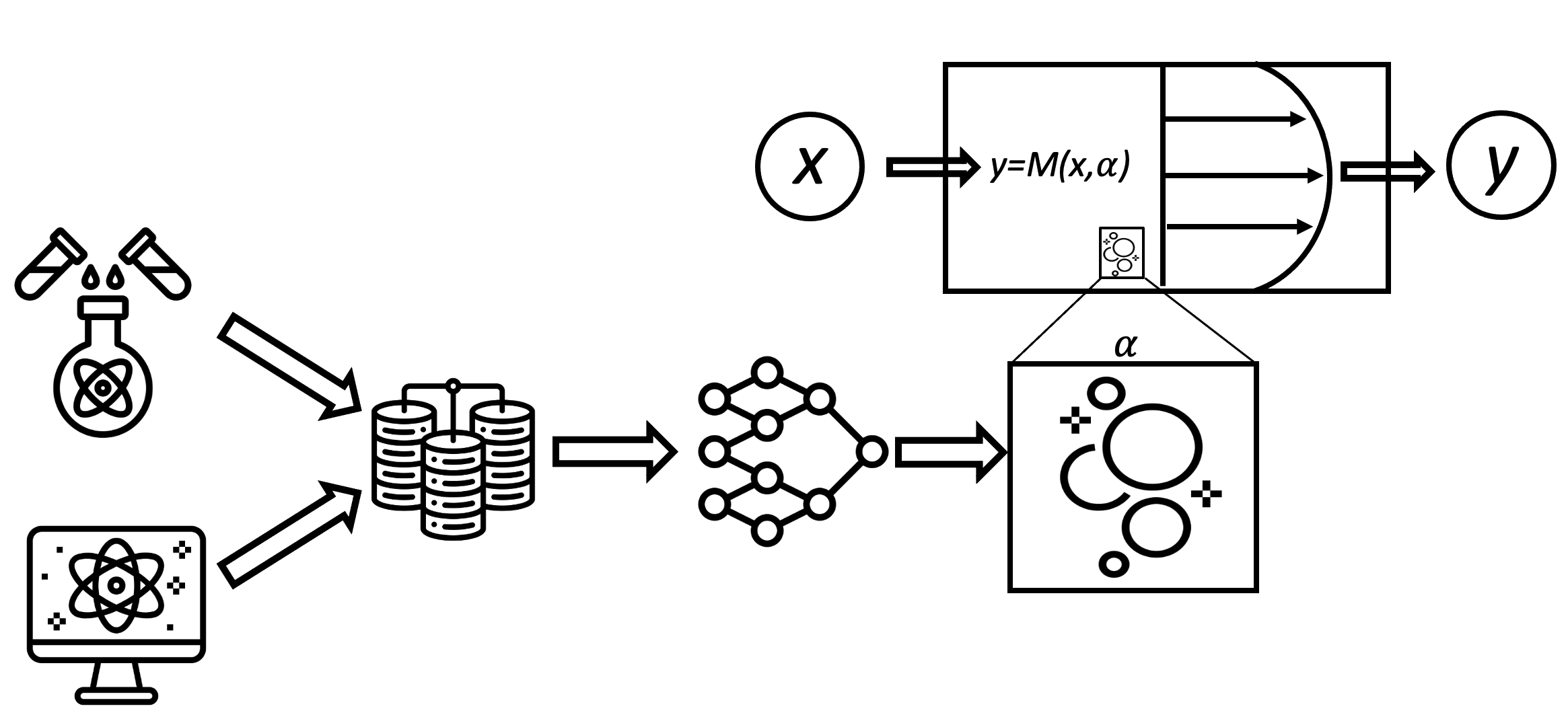}
    \caption{{\bf Example of machine learning applied to a case where partial knowledge is available about the underlying system}, illustrating a model (for instance a flow with complex rheology or a flow through a porous medium, top right of the picture) which depends on a set of known inputs $\mathbf{x}$ ({\it e.g.} geometry, boundary conditions, {\it etc.}) as well as on a set of hidden (unobservable) variables $\boldsymbol{\alpha}$ describing, {\it e.g.}, the fluid constitutive behavior. The latter may involve small-scale phenomena that can be difficult or impossible to describe. In such conditions, experimental or numerical data for observable quantities ({\it e.g.} velocity fields or stresses $\mathbf{y}$) can be used to infer the unknown field by training a machine learning model (here represented as a neural network, although other ML approaches are possible), subjected to physical constraints ({\it e.g.} positivity, symmetries or invariances). The whole process allows, on the one hand, to train a data-driven closure model for the hidden variables $\boldsymbol{\alpha}$ and, on the other hand, to gain \emph{a-posteriori} physical knowledge \textcolor{black}{{of}} the fluid constitutive properties.}
    \label{fig:knowsomething}
\end{figure}

In contrast to systems with well-understood governing equations, in some scientific problems we only have access to partial knowledge of the underlying mechanisms, as seen in complex materials ({\it e.g.}, composites or textured materials) and certain fluids ({\it e.g.}, granular or multiphase flows). These systems can exhibit simple microscopic behaviors yet result in complex macroscopic phenomena, such as the spread of infectious diseases or ``active turbulence'' in biological matter~\cite{souza2022relating,alert2022active}. In these cases, inductive biases, like frame invariance or symmetry constraints, can be incorporated into ML models to improve the discovery process~\cite{cranmer_et_al,liu2021physics}. Physical constraints ({\it e.g.}, thermodynamics) help create more generalizable models. For example, Moya {\it et al.}~\cite{moya_et_al} proposed a neural network constrained by known thermodynamic properties, enabling broader applications across physical systems. It is also important to highlight digital twins, where only some data are available, and they combine data-driven aspects with simulations while preserving generalization properties~\cite{willcox_digital}. Another example of taking advantage of known physical properties of the system is embedding symmetries in autoencoders, intending to develop reduced-order models (ROMs) in physical systems invariant to input transformations~\cite{kneer_et_al,otto_et_al}. These data-driven and physics-informed techniques are crucial for scientific discovery;  for instance, ML is helping to derive constitutive laws for materials with complex rheologies. De Lorenzis and collaborators~\cite{flaschel2023automated} introduced a hybrid framework (EUCLID) to learn constitutive equations for hyperelastic solids, 
and the SpaRTA framework for data-driven turbulence modeling \cite{schmelzer2020discovery} has been adapted to model
elastic solids \cite{wang2022establish}. Such approaches have also been used to infer constitutive equations from crystal structures and rheology of complex fluids~\cite{mahmoudabadbozchelou2022digital} (see Figure~\ref{fig:knowsomething}).

In the context of Life Sciences, structural biology is a field where ML has made significant advances despite partial knowledge of the phenomena. AlphaFold~\cite{jumper2021highly}, for instance, folds proteins into their three-dimensional (3D) native form from a one-dimensional(1D) amino acid sequence using deep learning with embedded biases like multiple sequence alignment (MSA) and 3D equivariance. AlphaFold has also spurred innovations in ML, including single-sequence methods like ESMfold~\cite{lin2023evolutionary} and generative models for protein design~\cite{Madani36702895,Dauparas36108050}. Diffusion models are also being used to generate protein-backbone structures and molecular ensembles, bypassing the need for simulations~\cite{RFdiffusion,Chroma}. Furthermore, generative models have also been used to discover physical models, integrating prior knowledge with data to generalize to new scenarios~\cite{cornelio2023combining}. Transformers, for instance, are now used in Chemistry to complete chemical reactions and predict reaction yields~\cite{irwin_et_al}. When it comes to quantum systems, deep reinforcement learning (DRL) has enabled new approaches to manipulate quantum states, providing insights into quantum mechanics~\cite{melnikov2018active}. Additionally, ML aids in reducing noise in quantum-computing systems, while quantum computing improves ML performance~\cite{melnikov2023quantum}. In climate modeling, ML has developed new LES models to ensure stable long-term forecasts~\cite{frezat2022posteriori}. ML has also been shown to enhance traditional weather-prediction systems~\cite{molina2023review}. 

\section*{Machine-learning-driven scientific discovery when little information is available}\label{sec:nothing}

There are numerous phenomena across scientific disciplines whose origins and underlying principles remain elusive. In these cases, the absence of well-established governing equations or foundational physical models makes it difficult to fully capture and understand their critical dynamics. For example, neuroscience has no first-principle equations, as there are no known conservation laws or symmetries to derive generalizable differential equations. Even with equations describing molecules or cells, the brain's complexity makes full-scale simulation unfeasible. Despite this, advances in neural data acquisition, such as large-scale neural recordings and connectomics, produce unprecedented datasets, promising a new era of ML-driven discovery in neuroscience and behavior~\cite{steinmetz_distributed_2019,yao2023whole}. Although full brain simulations remain beyond reach, data-driven models can replicate key input-output relationships, generating predictions vital for discovery. In neuroscience, perturbation experiments (such as activating neuron populations during visual tasks) provide insights into visual perception. Data-driven models can generate testable hypotheses and refine their predictions iteratively through experiments. ML also synthesizes diverse experimental data, such as the MICrONS dataset, which links functional imaging with structural reconstructions of cortical neurons~\cite{microns2021functional}.

\begin{figure}[ht]
\centering 
\includegraphics[width=0.75\textwidth]{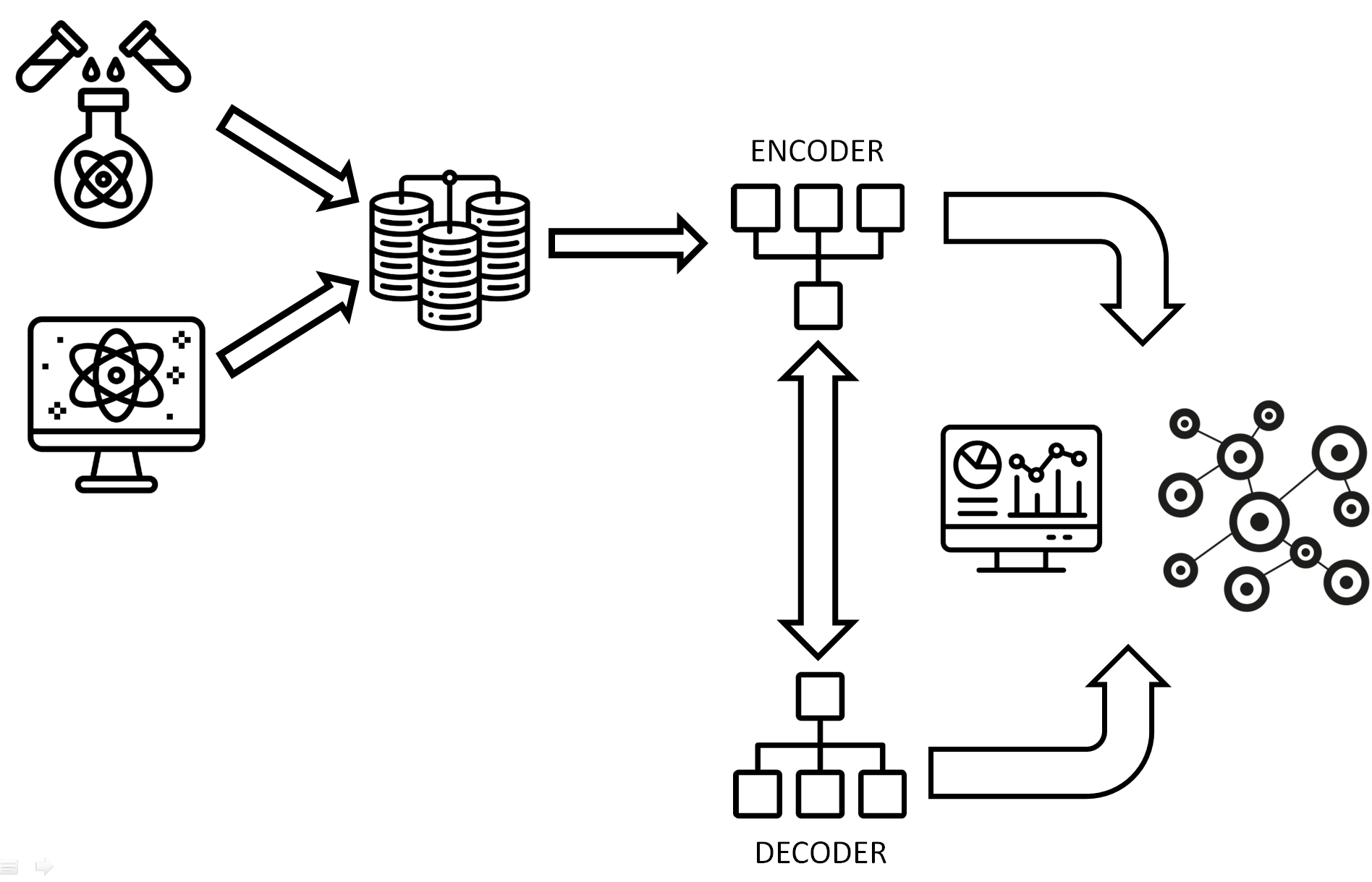}
   \caption{{\bf Schematic representation of a model (for instance, the observed symptoms of an unknown or complex disease within a population, or observed opinion dynamics within a social network) where the behavior as observed in data depends on an unknown dynamic or causal structure.} The observed behavior or dynamics might occur on several different spatial and temporal scales, and the observed data might reflect more or \textcolor{black}{{fewer}} aspects of the underlying system. In such conditions, {\it representation-learning} methods can be employed to distil out an explanation of the observed data in the form of a system of ODEs or as a causal-graph representation.}
   \label{fig:knownothing}
\end{figure}

ML can learn dynamics  where no a-priori knowledge exists, as seen in the Hodgkin--Huxley equations modeling neural dynamics~\cite{hheq}. Approaches like SINDy (sparse identification of nonlinear dynamics)~\cite{Brunton2016pnas}, genetic algorithms~\cite{chen2022symbolic} and reinforcement learning~\cite{du2022discover} can help uncover system dynamics. Neural ordinary differential equations (NODEs)~\cite{chen2018neural} are another method for modeling continuous dynamics, although they often lack scientific insight. Hybrid models relying on transformers~\cite{becker2023predicting} or SINDy-inspired architectures~\cite{sahoo2018learning} aim to bridge this gap. Interpretable-ML models can also discover biomarkers for diseases or predict treatment outcomes from patient data~\cite{qiu2020development}. In systems lacking governing equations, interventional data like CRISPR-Ko experiments in single-cell biology~\cite{ji2021machine} are enabling ML to identify some of the underlying mechanisms. Automated setups are advancing\cite{macleod2022self}, leading to research in designing experiments for system identification, especially for non-linear models like NODEs~\cite{du2020model}. Representation-learning techniques, such as variational autoencoders~\cite{bengio2013representation}, are essential for simplifying the characterization of complex systems by identifying latent spaces that reduce dimensionality and reveal causal structures~\cite{scholkopf2021toward}. Causality and its application to dynamical systems have gained prominence~\cite{causality_review}, especially in Earth sciences~\cite{runge2019inferring} and molecular biology~\cite{lobentanzer2024molecular}, with some studies using invariance from heterogeneous experiments as a signal to identify causal ODEs~\cite{pfister2019learning}. Learning structured latent spaces is of crucial importance since it provides an effective coordinate system in which the dynamics have a simple representation, which is a key requirement for generalization and interpretability~\cite{Champion2019pnas}. In Figure~\ref{fig:knownothing} we provide a schematic representation of the identification of an underlying causal structure from observations where the variables of interest are not directly observed.

Recent research has also explored improving diffusion models and solvers~\cite{lu2022dpm}, which are useful for generating state-of-the-art results in areas like image generation~\cite{rombach2022high}, protein modeling~\cite{watson2023novo} and materials science~\cite{zeni2023mattergen}. Despite their success, large pre-trained models provide limited scientific insights~\cite{nichani2024transformers}, and therefore constitute an opportunity for future research. Machine learning has also enhanced data collection and processing in systems with indirect or incomplete measurements. ML imputation and generative modeling can complete time-series data, improving downstream applications~\cite{vetter2023generating}. Furthermore, computer-vision techniques have also automated previously manual tasks, such as segmentation in microscopy, enabling large-scale analysis of cell populations~\cite{kirillov2023segment}. In ethology, ML has transformed video data into animal kinematics and poses, linking behavior to underlying neural computations~\cite{mathis_deeplabcut_2018}. Notably, large language models (LLMs) and scientific foundation models (SFMs) are opening new avenues for extracting scientific insights directly from data. Genome-scale language models  (GenSLMs), for instance, are helping to learn the evolutionary landscape of SARS-CoV-2 genomes~\cite{zvyagin2023genslms}, while LLMs are enhancing neuroscience research by integrating diverse datasets and summarizing insights across isolated subfields~\cite{bzdok2024data}.

In sum, it is hard to overstate the ongoing impact of machine learning as a critical tool that, when used in conjunction with other approaches ({\it e.g.} experiments, causality analysis, development of an adequate coordinate system, {\it etc.}), catalyzes advances in scientific fields where no information on the phenomenon under study is available.

\section*{The drawbacks, limitations and challenges of machine learning for scientific discovery}

Despite the significant advances that ML has brought to scientific discovery, there are key areas that need to be addressed to accelerate its contributions and realize its full potential while minimizing some clear risks, as discussed below:
\begin{itemize}
    \item Data-related challenges: The success of ML in scientific discovery relies on large high-quality, structured datasets, but scientific data is often incomplete, noisy, or imbalanced, leading to biased models. Unstructured data, especially in fields like biology, chemistry, and geology, complicate the use of ML applications, since these algorithms are not inherently designed to handle such data. However, it was recently shown that foundation models could help for prediction tasks in small general datasets~\cite{TabPFN}. Furthermore, the absence of labeled data complicates the use of supervised-ML techniques~\cite{j31}.
    \item Bias and ethical issues: ML models are prone to data bias, distorting results and hindering scientific discovery. Bias in data collection or model training can reinforce existing assumptions instead of revealing novel insights~\cite{j32}. This is especially concerning in fields such as drug discovery, where existing models might overlook demographic groups, making the discoveries less generalizable~\cite{j35}. Ethical concerns, especially in medicine, highlight risks in applying ML to decisions affecting human life~\cite{j37}.
    \item Explainability and interpretability: ML models, particularly those based on deep learning, often function as ``black boxes'', making accurate predictions without revealing their decision-making processes~\cite{vinuesa_interp}. This lack of transparency is problematic in fields like drug discovery, where understanding why a model predicts a certain interaction is crucial~\cite{j40}. Advances in explainable AI (XAI) have improved interpretability in areas like materials science~\cite{j41}, chemistry~\cite{j43}, and medicine~\cite{j46}.
    \item Overfitting and generalization: ML models often overfit their training data, performing poorly on unseen data, a fact that limits their ability to generalize. In scientific contexts, this can produce misleading results and fail to capture complex, nonlinear relationships, as seen in chemistry, biology, and astronomy~\cite{j47,j51,j52}. Overfitting restricts the potential of ML to develop universal theories essential for broad scientific understanding~\cite{j53}.
\end{itemize}

Improving data quality and diversity is critical for ML models to generalize across disciplines. Initiatives like the Open Reaction Database~\cite{j55} and the Crystallography Open Database~\cite{j56} aim to enhance data management under the FAIR (findability, accessibility, interoperability and reusability) principles~\cite{j57,barba}. Incorporating bias detection and fairness-aware algorithms can reduce biased data impacts~\cite{j59}. XAI methods are being developed to improve the transparency of ML models, particularly in healthcare applications like brain-tumor segmentation~\cite{j61}. Hybrid approaches that combine ML with physics-based models yield promising results by ensuring that the predictions adhere to known scientific principles~\cite{j38,j67}. It is also important to note that ethical frameworks are also needed to ensure fairness, transparency, and accountability, especially in medicine~\cite{j70,j72}.

While AI has shown great potential in scientific discovery, significant challenges remain. Issues related to data quality, bias, interpretability, and overfitting must be addressed to harness the full potential of AI in advancing science. By improving data access, enhancing model transparency, and developing hybrid models, the scientific community can overcome these obstacles and drive meaningful, trustworthy discoveries using AI which may play an instrumental role in areas as far-reaching as gravitational waves~\cite{j74}, space exploration~\cite{j75} or even the discovery of extraterrestrial life~\cite{j77}.

\section*{Conclusions and outlook}\label{sec:conclusions}

Karl Popper famously stated in {\it The Open Universe: An Argument for Indeterminism from the Postscript to The Logic of Scientific Discovery} that ``science can be described as the art of systematic oversimplification''~\cite{j78} and this remains true due to the historical limitations in processing and analyzing vast amounts of data. As a result, scientists have been forced to simplify their objects of study by focusing on isolated phenomena or simple models. While simplification has benefits, for example in the teaching and the systematization of science, oversimplification carries significant risks, as essential information and key aspects of problems may be lost in the process. For example, in ecological research, studies often focus on individual species interactions without considering the broader dynamics of ecosystems or other effects, such as climate change~\cite{j79}.

ML methods have already enabled several scientific and technological advancements, from solving image classification to winning at Go. This article highlights how modern data-driven methods are enabling breakthroughs in scientific discovery, focusing on state-of-the-art techniques that push beyond previous limitations. We categorize these approaches by how much knowledge about the underlying mechanisms is available, ranging from well-understood systems to those with unknown governing principles. As shown in Table~\ref{tab:summary}, ML's potential spans a wide spectrum of applications, including discovering physical laws, evaluating complex systems, inferring unknown behaviors, and uncovering multiscale principles. These advances apply across fields like Physics, Mathematics, Chemistry, and Life Sciences, promising faster scientific progress than ever before. However, using ML for scientific discovery brings challenges. A key advantage of ML is its ability to model complex systems, but this often requires extensive training data. In areas like astrophysics, rare diseases, or new pharmaceuticals, data scarcity is a frequent obstacle. Fortunately, complementary ML methods can either generate needed data or bypass large datasets through techniques like self-supervised learning. Even when ML does not directly result in discoveries, it can facilitate breakthroughs where data are limited. Validation is another challenge, especially in cases where the governing principles are unknown. But ML-driven discoveries can be confirmed using traditional scientific methods, such as hypothesis testing, observational confirmation and benchmarking. Additionally, ML models often function as ``black boxes'', making it hard to derive formal knowledge from their results. This is a significant issue for science, where understanding is key. However, explainable- and interpretable-ML methods offer solutions, helping to achieve discoveries in the context of established scientific principles. Despite these challenges, ML is becoming an essential tool across disciplines, and its continued evolution promises even more opportunities for scientific discovery. With further refinement, ML techniques are likely to address the current limitations, enabling even greater advancements.
\begin{table}
\centering 
\includegraphics[width=0.85\textwidth]{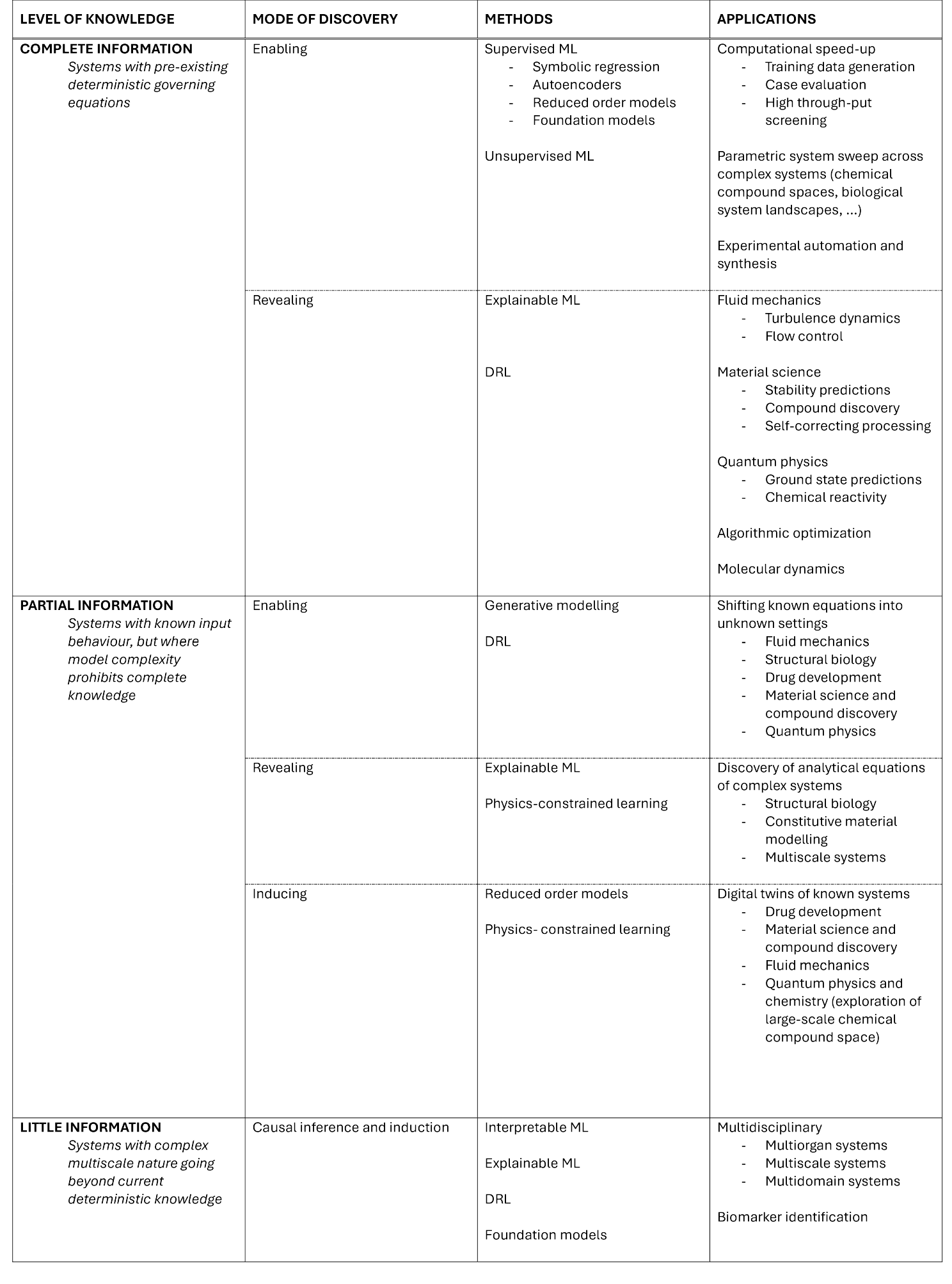}
   \caption{{\bf Summarizing overview of the opportunities for machine learning in scientific discovery.} Based on the differentiation presented in our work, the level of prior, deterministic knowledge ({\it left}) can be used to differentiate methods ({\it second right}) and applications ({\it right}) across which scientific advancements can be made by means of dedicated AI systems. This also allows for various modes of discovery ({\it second left}), ranging from cases where machine learning enables discovery by allowing for efficient computational usage, parametric sweeps, etc., to cases where machine learning is used to causally infer underlying mechanistic behaviours in complex multidisciplinary systems.}
   \label{tab:summary}
\end{table}

The recent Nobel Prizes in Physics~\cite{np} and Chemistry~\cite{nc} underscore the transformative impact of machine learning, demonstrating its ability to revolutionise scientific discovery. By providing cutting-edge tools for data analysis, ML is accelerating breakthroughs across multiple fields, deepening our understanding and tackling complex challenges in ways previously unimaginable. %AI now allows us, for the first time in history, to capture the true complexity of reality. 
This capability is crucial, as only by embracing this complexity can we connect the dots in science, grasp the intricate interdependencies of natural systems, and facilitate breakthrough, unpredictable discoveries.
%Furthermore, machine learning can analyze massive data sets and uncover patterns and connections that would otherwise remain hidden. Only by embracing this complexity can we connect the dots in science, understand the intricate interdependencies of natural systems, and make breakthrough, unpredictable discoveries (such as identifying hidden molecular interactions in drug discovery or predicting climate anomalies from multifactorial data inputs~\cite{j80,j81}). 
The real challenge -- and opportunity -- of AI-driven scientific discovery lies in moving beyond solving narrow, well-defined problems to tackling complex, open-ended questions. This shift calls for the development of artificial general intelligence (AGI) or even artificial superintelligence (ASI), which possesses the potential to engage in multifaceted problem-solving across diverse domains. AGI for scientific discovery could connect seemingly disparate areas of knowledge, leading to breakthroughs that single-focused AI cannot achieve. For instance, an AGI model could integrate insights from pharmacology, neuroscience, and data science to not only propose novel drug candidates but also predict their interactions and effects on complex biological systems. This capacity for interdisciplinary connections would facilitate original contributions to scientific knowledge, allowing for innovative solutions to intricate challenges. Ultimately, harnessing AGI's multifaceted problem-solving abilities presents a transformative opportunity to revolutionize how we approach scientific discovery and address some of humanity's most pressing challenges.

% \begin{spacing}{.7}
% \setlength{\bibsep}{1.pt}
%\bibliographystyle{abbrvnat}
\bibliography{serc_bib,mybib}
% \end{spacing}

%\subsection{Key references}

%\begin{itemize}
%    \item~\citealp{alpha_fold}
%\end{itemize}

%\noindent \textbf{For Reviews only, highlighted references (optional)} Please select 5–-10 key references and provide a single sentence for each, highlighting the significance of the work.

\section*{Acknowledgements}
The following researchers are acknowledged for helpful discussions during the preparation of this article: Frida Bender, Annica Ekman, Inga Koszalka, Romit Maulik, Henrik Nielsen, Gunilla Svensson, Bj\"orn Wallner. RV and HA acknowledge SeRC and Digital Futures for funding the workshop that initiated this work. RV acknowledges financial support from ERC grant no. `2021-CoG-101043998, DEEPCONTROL'. DM acknowledges financial support from ERC grant no. 2022-StG-101075494, MultiPRESS. Views and opinions expressed are, however, those of the author(s) only and do not necessarily reflect those of the European Union or the European Research Council. Neither the European Union nor the granting authority can be held responsible. AE was funded by the Vetenskapsr\r{a}det Grant No. 2021-03979 and the Knut and Alice Wallenberg Foundation and by SeRC. SLB acknowledges funding support from the US National Science Foundation AI Institute in Dynamic Systems
(grant number 2112085) and from The Boeing Company.

\section*{Author contributions}
RV and HA initiated the idea for this article following a workshop celebrated in November 2022 at KTH. All the authors contributed equally to the rest of this work.

\section*{Competing interests}
The authors declare no competing interests. 

\section*{Publisher’s note}
Springer Nature remains neutral with regard to jurisdictional claims in published maps and institutional affiliations.

%\section*{Supplementary information (optional)}
%If your article requires supplementary information, please include these files for peer-review. Please note that supplementary information will not be edited.

%\clearpage
%\newpage
%\section{Box 1 (Optional)}
%This is a Box, which can contain a figure, and which should have no more than 300 words of text.

%\section{Figures}

%\clearpage
%\newpage
%\section{Tables}

\end{document}